\documentclass[11pt]{article}

\usepackage[final]{neurips_2024}

\makeatletter
\renewcommand{\@notice}{}
\makeatother

\usepackage[most]{tcolorbox}
\usepackage{xcolor}
\usepackage[table]{xcolor}

\definecolor{abstractbg}{RGB}{241, 246, 254} 

\newtcolorbox{abstractbox}{
  colback=abstractbg,      
  colframe=white,      
  boxrule=0pt,         
  arc=6pt,             
  left=10pt,
  right=10pt,
  top=10pt,
  bottom=10pt
}

\usepackage{float}

\usepackage[utf8]{inputenc}
\usepackage{booktabs}
\usepackage{multirow}
\usepackage{graphicx}
\usepackage{geometry}

\usepackage{tabularx} 

\usepackage{enumitem}

\usepackage{amsmath,amssymb}

\usepackage{graphicx}
\usepackage{booktabs}
\usepackage{multirow}
\usepackage{caption}
\usepackage{subcaption}

\usepackage{booktabs}
\usepackage{tabularx}
\usepackage{multirow}

\usepackage[table]{xcolor} 

\usepackage{array} 

\usepackage[dvipsnames]{xcolor}
\usepackage[hidelinks]{hyperref}

\usepackage{listings}
\usepackage{courier}
\usepackage{enumitem}
\setlist[itemize]{noitemsep, topsep=4pt}
\setlist[enumerate]{noitemsep, topsep=4pt}

\usepackage{titlesec}
\titleformat{\section}{\large\bfseries}{\thesection}{0.6em}{}
\titleformat{\subsection}{\normalsize\bfseries}{\thesubsection}{0.6em}{}

\title{OmegaUse: Building a General-Purpose GUI Agent for Autonomous Task Execution}
\author{
Le Zhang$^*$, Yixiong Xiao$^*$, Xinjiang Lu$^*$, Jingjia Cao$^*$, Yusai Zhao$^*$, Jingbo Zhou${^{*\dagger}}$, \\
\textbf{Lang An, Zikan Feng, Wanxiang Sha, Yu Shi, Congxi Xiao, Jian Xiong, Yankai Zhang,}\\
\textbf{Hua Wu$^{\dagger}$, Haifeng Wang$^{\dagger}$}\\
Baidu Frontier Research Department\\
*Equal contribution; 
$^{\dagger}$Contact authors:\{zhoujingbo, wu\_hua, wanghaifeng\}@baidu.com
}
\date{}

\begin{document}

\maketitle

\begin{abstractbox}
\begin{abstract}

Graphical User Interface (GUI) agents show great potential for enabling foundation models to complete real-world tasks, revolutionizing human–computer interaction and improving human productivity. In this report, we present \textbf{OmegaUse}, a general-purpose GUI agent model for autonomous task execution on both mobile and desktop platforms, supporting computer--use and phone-use scenarios. Building an effective GUI agent model relies on two factors: (1) high-quality data and (2) effective training methods. To address these, we introduce a carefully engineered data-construction pipeline and a decoupled training paradigm.
For data construction, we leverage rigorously curated open-source datasets and introduce a novel automated synthesis framework that integrates bottom-up autonomous exploration with top-down taxonomy-guided generation to create high-fidelity synthetic data. For training, to better leverage these data, we adopt a two-stage strategy: Supervised Fine-Tuning (SFT) to establish fundamental interaction syntax, followed by Group Relative Policy Optimization (GRPO) to improve spatial grounding and sequential planning. To balance computational efficiency with agentic reasoning capacity, OmegaUse is built on a Mixture-of-Experts (MoE) backbone. To evaluate cross-terminal capabilities in an offline setting, we introduce \textbf{OS-Nav}, a benchmark suite spanning multiple operating systems: \textbf{ChiM-Nav}, targeting Chinese Android mobile environments, and \textbf{Ubu-Nav}, focusing on routine desktop interactions on Ubuntu. Extensive experiments show that OmegaUse is highly competitive across established GUI benchmarks, achieving a state-of-the-art (SOTA)  score of \textbf{96.3\%}  on ScreenSpot-V2 and a leading \textbf{79.1\%}  step success rate on AndroidControl. OmegaUse also performs strongly on OS-Nav, reaching \textbf{74.24\%} step success on ChiM-Nav and  \textbf{55.9\%} average success on Ubu-Nav. 

\end{abstract}
\end{abstractbox}


\section{Introduction}

\begin{figure}[t]
    \centering
    \includegraphics[width=0.6\textwidth]{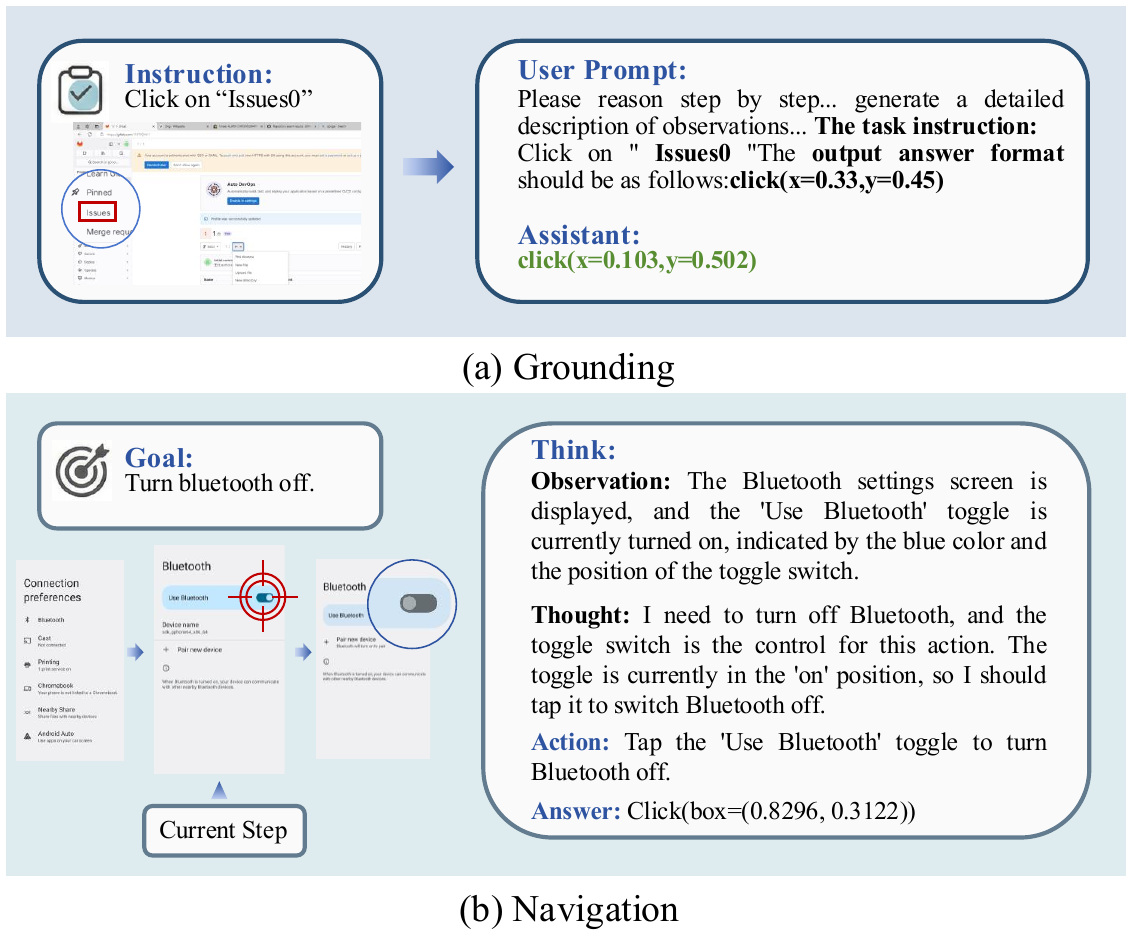} 
    \caption{An overview of OmegaUse's core capabilities in realistic GUI scenarios.}
    \label{fig:omniuse_illustration}
\end{figure}

GUI agents have recently emerged as a transformative frontier for multimodal interaction, enabling artificial intelligence to navigate digital environments ranging from mobile applications to desktop software in a manner analogous to human users~\cite{hong2024cogagent,ui_tars_2025,wu2024osatlas}. By perceiving screen states through screenshots and executing atomic actions such as clicking, typing, and scrolling, these agents aim to bridge the gap between high-level user intent and complex operational sequences~\cite{ hong2024cogagent,cheng2024seeclick}, as illustrated in Figure~\ref{fig:omniuse_illustration}. Depending on the target platform, they are often referred to as computer-use, phone-use, or browser-use agents.


Despite significant progress, current GUI agents still face critical bottlenecks in performance, training-data quality, and the lack of comprehensive evaluation across diverse digital ecosystems. To address these challenges, we present \textbf{OmegaUse}, a general-purpose GUI agent model\footnote{While the term “GUI agent” typically refers to the full system that interacts with a digital environment (e.g., including external tools), our work focuses on end-to-end model training. We treat GUI agentic capability as a high-level policy learned via a dedicated model-based approach.} designed for autonomous task execution. We name the agent OmegaUse to reflect its unified support for both computer-use and phone-use scenarios across diverse platforms. OmegaUse is built on a Mixture-of-Experts (MoE) backbone. Compared with compact dense models (e.g., 7B or 72B)~\cite{ui_tars_2025, gu2025ui, bai2025qwen25vl}, this design preserves the reasoning capacity of large-parameter models while activating only a subset of parameters, enabling superior performance with substantially reduced computational overhead.

We acknowledge that data quality is a primary determinant of GUI agent model performance, as noisy training signals can substantially degrade both spatial perception and decision-making. In grounding tasks, labels automatically derived from HTML or Accessibility (A11y) trees often suffer from rendering offsets, leading to misaligned bounding boxes and ambiguous textual descriptions~\cite{cheng2024seeclick, gu2025ui}. Moreover, existing navigation datasets frequently contain inconsistencies, such as incorrect execution trajectories and excessive redundant actions, that provide weak or incoherent supervision for long-horizon planning~\cite{gu2025ui, li2024effects}.

To address these issues, we develop a high-quality training-corpus construction pipeline. For grounding, we apply a stringent filtering procedure to improve label precision. For navigation, we propose a novel hierarchical synthesis framework that integrates three complementary data sources: (1) rigorously curated open-source datasets, (2) automatically synthesized trajectories generated by combining bottom-up autonomous exploration with top-down taxonomy-guided generation, and (3) high-fidelity cross-terminal expert demonstrations.

To effectively leverage the curated data, we propose a decoupled two-stage training paradigm. We first apply supervised fine-tuning (SFT) to establish foundational interaction syntax and basic task logic~\cite{hong2024cogagent, ui_tars_2025}. We then employ Group Relative Policy Optimization (GRPO) to refine spatial grounding and sequential planning~\cite{deepseek2025r1, shao2024deepseekmath, gu2025ui}. With specialized reward design, such as an \emph{Inside-of-Bounding-Box} reward for grounding and stepwise coordinate-based rewards for navigation, OmegaUse is encouraged to focus on precise interaction regions rather than ambiguous boundary pixels~\cite{gu2025ui, zhou2025guig1, tang2025guig2}.

Beyond architecture and training strategies, we observe that existing benchmarks may not fully capture an agent’s proficiency across diverse digital environments, such as Chinese mobile applications or multi-step desktop workflows. To complement current evaluation resources, we introduce \textbf{OS-Nav}\footnote{https://huggingface.co/datasets/baidu-frontier-research/OS-Nav}, a specialized offline benchmark comprising two sub-benchmarks across different operating systems: \textbf{ChiM-Nav}, focusing on Chinese Android mobile systems, and \textbf{Ubu-Nav}, targeting routine desktop interactions on Ubuntu. Both datasets provide expert-verified reasoning trajectories, enabling a more comprehensive evaluation of agent generalization and planning consistency.

Through extensive empirical evaluations, we demonstrate that OmegaUse consistently outperforms or remains competitive with SOTA GUI agents across multiple platforms. On standard grounding benchmarks, OmegaUse achieves a record \textbf{96.3\%} on ScreenSpot-V2. In navigation tasks, it reaches a leading \textbf{79.1\%} step success rate on AndroidControl and demonstrates robust interactive capabilities on AndroidWorld. Furthermore, on our proposed benchmarks, OmegaUse delivers superior performance with a \textbf{74.24\%} step success rate on ChiM-Nav and a \textbf{55.9\%} average success rate on Ubu-Nav. These results underscore the effectiveness of OmegaUse agent.

Our main contributions are summarized as follows:

\begin{itemize}[leftmargin=0.7cm]
    \item We introduce \textbf{OmegaUse}, a general-purpose GUI agent built on a parameter-efficient MoE architecture for autonomous task execution. OmegaUse is trained using a decoupled two-stage paradigm, and we present a holistic framework for building GUI agents that jointly addresses data construction and model training.

    \item We establish a high-quality data foundation for GUI agents. In particular, we propose a hierarchical navigation data construction pipeline featuring a novel automated synthesis framework that combines bottom-up autonomous exploration with top-down taxonomy-guided generation. This approach substantially reduces reliance on manual annotations while ensuring data diversity, coverage, and fidelity across platforms.

    \item To bridge the evaluation gap in specific digital environments,  we release \textbf{OS-Nav}, a specialized offline benchmark suite comprising \textbf{ChiM-Nav} for Chinese Android mobile ecosystems and \textbf{Ubu-Nav} for routine Ubuntu desktop system. OS-Nav enables rigorous assessment of cross-platform generalization and planning consistency.

    \item Extensive empirical evaluations demonstrate that OmegaUse achieves highly competitive performance across a wide range of GUI benchmarks, including state-of-the-art results on several tasks. Notably, OmegaUse attains a record \textbf{96.3\%} accuracy on ScreenSpot-V2 and a leading \textbf{79.1\%} step success rate on AndroidControl.
\end{itemize}

\section{Related Work}
Recent years have witnessed rapid progress in GUI agents, which are models that perceive GUI states (e.g., screenshots and/or structured UI representations) and execute actions (e.g., clicking, typing, and scrolling) to accomplish user goals. In this section, we review prior work along two main axes: (1) UI grounding and GUI perception; and (2) GUI agent architectures, including modular pipelines and native (end-to-end) agent models.

\subsection{UI Grounding and GUI Perception}
Accurate \emph{UI grounding}, which aligns natural language references with specific GUI elements on the screen, is widely recognized as a core bottleneck for GUI agents. A representative line of work focuses on grounding-centric models that localize UI elements directly from screenshots and instructions, while establishing standardized evaluations for cross-platform generalization. Early approaches typically relied on supervised learning over annotated screenshots, predicting click points or bounding boxes conditioned on natural language instructions~\cite{som_2023,qian2024visual,gou2024navigating}.

Representative efforts such as SeeClick~\cite{cheng2024seeclick} and subsequent grounding-oriented models~\cite{zhang2024mm1,qian2024visual,nguyen2024improved,lin2025showui} demonstrated the feasibility of instruction-conditioned UI localization, but also revealed strong sensitivity to screen resolution, layout diversity, and domain shift. To better characterize these challenges, several benchmarks have been proposed. ScreenSpot~\cite{cheng2024seeclick} introduced cross-platform grounding evaluation across mobile, web, and desktop interfaces, while ScreenSpot-V2~\cite{wu2024osatlas} improves upon ScreenSpot by revising and correcting its original annotations. Follow-up datasets such as ScreenSpot-Pro~\cite{li2025screenspot} further emphasize small targets and professional workflows. 

Subsequent strong baselines, often reused across later agent studies, include OS-Atlas~\cite{wu2024osatlas}, Aguvis~\cite{xu2024aguvis}, and UGround~\cite{qian2025uground}. Together, these works demonstrate that GUI grounding performance remains highly sensitive to resolution, layout diversity, and distribution shift. More recent work explores reinforcement-learning-style post-training for UI grounding, in which rewards directly reflect spatial correctness to improve generalization and reduce dependence on dense annotations. Examples include UI-R1~\cite{lu2025ui}, GUI-R1~\cite{luo2025gui}, InfiGUI-R1~\cite{liu2025infigui}, and coordinate-free grounding approaches such as GUI-Actor~\cite{wu2025gui}. Related variants investigate reward modeling and policy optimization strategies tailored to GUI grounding, including GUI-G2~\cite{tang2025guig2} and InfiGUI-G1~\cite{liu2025infigui}.

\subsection{GUI Agent Architectures: Modular Pipelines vs. Native Agents}
Early and many contemporary GUI agents adopt \emph{modular architectures}, decomposing the overall problem into separate components for perception, planning, memory, and execution. Agent-S~\cite{agashe2024agent},  Agent-S3~\cite{gonzalez2025unreasonable} and Cradle~\cite{tan2024cradle} exemplify framework-centric designs that leverage a strong foundation model for planning and reflection, while relying on explicit modules such as prompted planners, memory buffers, verifiers, and tool wrappers to improve controllability and interpretability. Mobile-Agent~\cite{wang2024mobile, wang2024mobilev2,ye2025mobilev3} follows a similar decomposition for mobile environments, using vision-based perception to reduce reliance on platform metadata. OS-Symphony~\cite{yang2026symphony} and GTA1~\cite{yang2025gta1} both advance computer-using agent frameworks by improving robustness and generalization for GUI-based tasks through careful system design and enhanced inference-time scaling.
A common pattern in these systems is to incorporate strong grounding models and optionally UI parsers, such as OmniParser~\cite{wan2024omniparser}, to obtain structured UI representations for downstream planning. However, modular pipelines are prone to error accumulation across components and often require extensive hand engineering to support diverse applications and long-horizon tasks.

In contrast, recent work has increasingly shifted toward \emph{native} or \emph{end-to-end GUI agents}, which unify perception, reasoning, and action within a single model. AutoWebGLM~\cite{lai2024autowebglm} and UI-TARS~\cite{ui_tars_2025} frame this shift as analogous to end-to-end tool-using agents, arguing that unified policies can more effectively leverage large-scale data and reinforcement-learning signals. AutoGLM~\cite{liu2024autoglm} introduces an intermediate interface to decouple planning from grounding and proposes a progressive, self-evolving online curriculum reinforcement-learning framework for web and mobile GUI control. UI-TARS-2~\cite{ui_tars2_2025} further emphasizes \emph{multi-turn reinforcement learning} as a key driver of performance gains, enabling agents to optimize long-horizon behavior and recover from intermediate errors. AgentCPM-GUI~\cite{zhang2025agentcpm} targets efficient on-device mobile GUI interaction by introducing a compact action space and a three-stage training pipeline. Step-GUI~\cite{yan2025step} proposes a self-evolving training pipeline and couples it with a hierarchical GUI-MCP protocol to enable standardized, privacy-preserving execution across heterogeneous devices.
OpenCUA~\cite{wang2025opencua} provides open foundations for computer-use agents, including datasets, evaluation protocols, and strong baselines. Mano~\cite{fu2025mano} investigates training strategies and system designs for general computer use, including iterative improvement and evaluation-oriented components that bridge framework-based approaches and end-to-end policy learning. UI-Venus~\cite{gu2025ui} further highlights the central role of data quality and trajectory curation in driving performance gains. MAI-UI~\cite{zhou2025mai} explicitly emphasizes deployment considerations, including agent–user interactive operation and MCP-augmented tool use. Although these methods enable implicit planning and memory to emerge from multi-step trajectory training, they also introduce challenges in training stability and environment scalability.


\section{Methodology}

\begin{figure}[t]
    \centering
    \includegraphics[width=\textwidth]{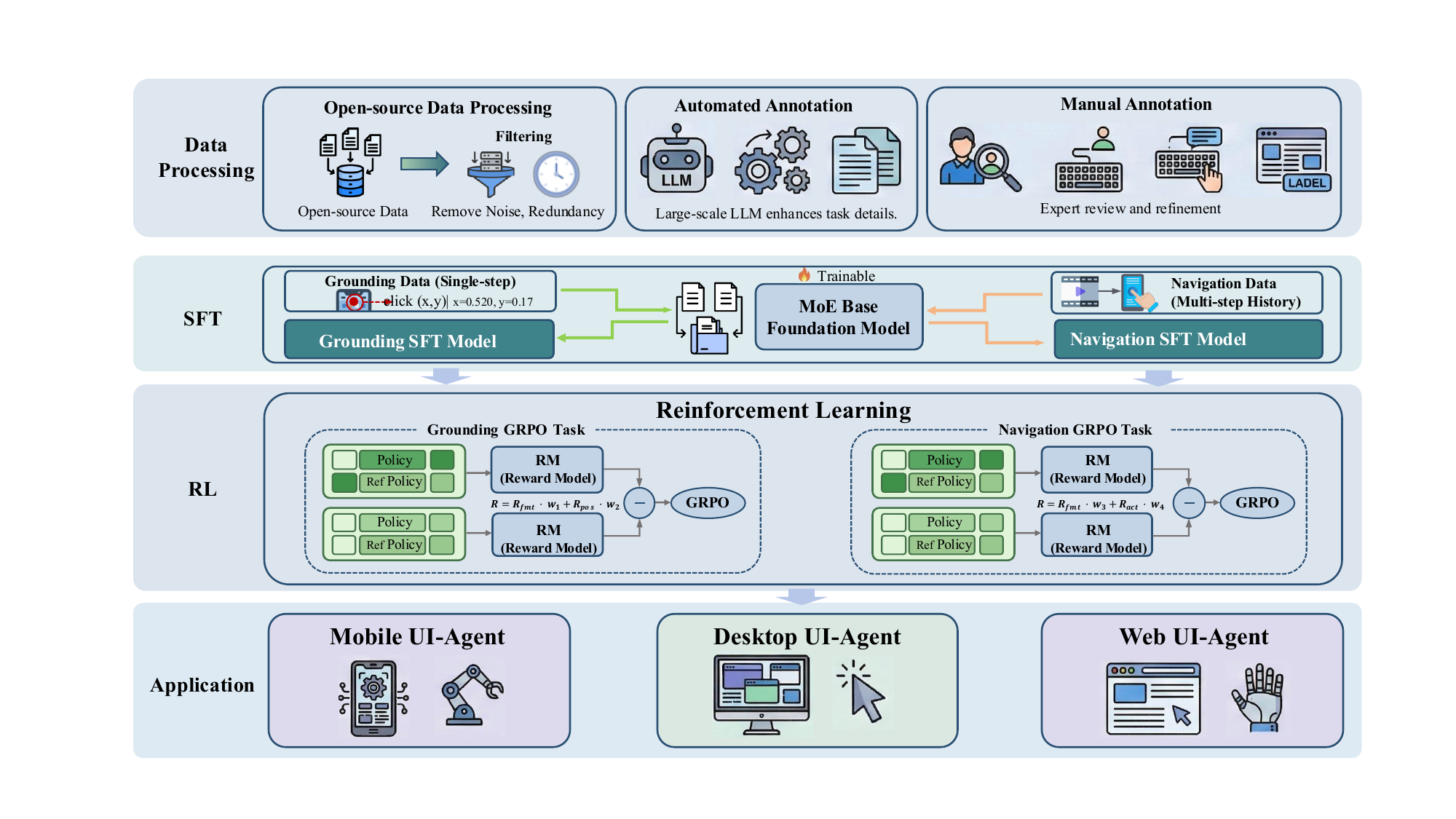} 
    \caption{The overall architecture of the OmegaUse framework. The pipeline proceeds through four distinct layers: (1) a hybrid data processing stage integrating automated LLM-assisted annotation and human-in-the-loop refinement; (2) SFT of an MoE foundation model; (3) decoupled RL using GRPO with tailored rewards for grounding and navigation tasks; and (4) final deployment of the optimized agents across diverse application environments.}
    \label{fig:omniuse_framework}
\end{figure}

Our training paradigm uses a decoupled design with two specialized models: (i) a grounding model for high-precision visual perception and (ii) a navigation model for sequential decision-making. Figure~\ref{fig:omniuse_framework} illustrates the overall framework architecture. This separation enables targeted optimization and reduces interference between low-level spatial grounding and high-level reasoning.

\subsection{{OmegaUse-G: Foundation of Visual Perception}}

The grounding model is designed to map textual queries to precise spatial coordinates on the UI. We first describe the data construction process for the grounding model and then present the corresponding training strategy.

\subsubsection{Grounding Data Pipeline}

We aggregated a diverse GUI grounding corpus by consolidating six publicly available datasets: Aguvis~\cite{xu2024aguvis}, UI RefExp~\cite{bai2021uibert}, Widget Captioning~\cite{li2020widget}, SeeClick~\cite{cheng2024seeclick}, Uground~\cite{qian2025uground}, and OS-Atlas~\cite{wu2024osatlas}. As summarized in Table 1, these sources provide a comprehensive coverage of mobile, web, and desktop interfaces. The combined raw pool contains approximately 1.66 million instances.

\begin{table}[htbp]
\centering
\caption{Statistics of the GUI grounding datasets used in our study. The raw pool of 1.66M instances was distilled into a 111k high-quality training set.}
\label{tab:data_summary}
\begin{tabular}{lcc}
\toprule
\textbf{Dataset} & \textbf{Platform} & \textbf{Raw Samples} \\
\midrule
Aguvis & Mobile & 110k \\
UI RefExp & Mobile & 16k \\
Widget Captioning & Mobile & 40k \\
SeeClick & Web & 250k \\
Uground & Web & 750k \\
OS-Atlas & Desktop & 490k \\
\midrule
\textbf{Total Raw Pool} & - & \textbf{1.66M} \\
\textbf{Final Sampled Set} & \textbf{Mixed} & \textbf{111k} \\
\bottomrule
\end{tabular}
\end{table}


Despite the large scale of existing open-source datasets, we observe that nearly 40\% of raw instances contain substantial noise, including misaligned bounding boxes and ambiguous textual prompts. These issues are particularly prevalent in datasets whose labels are automatically extracted from HTML or accessibility trees, where rendering offsets frequently introduce spatial inaccuracies. Prior studies~\cite{gu2025ui} have shown that data quality critically affects grounding performance, and fully automated filtering methods often struggle to reliably identify high-quality examples.

To address these bottlenecks, we employed a manual inspection and correction pipeline. 
We first eliminate redundant or overly simplistic samples, followed by downsampling to retain 300K instances. Subsequently, we manually realign shifted bounding boxes and rephrase ambiguous or meaningless instructions to enforce a precise one-to-one correspondence between visual elements and their textual descriptions. Besides, samples containing blurred images or inherently ambiguous instructions are strictly filtered out. This rigorous refinement process yields a curated dataset of 111K high-quality samples, ensuring that the model is trained with reliable supervision signals.

\subsubsection{Two-Stage Grounding Training}

To optimize the spatial reasoning and localization precision of our model, we adopt a hierarchical training paradigm using the manually refined grounding dataset. We partition the dataset into a transition from foundational coordinate formatting to high-precision reinforcement refinement.

\noindent\textbf{(1) Policy Initialization (SFT):} In the first stage, we perform SFT to establish the fundamental capability of the model to interpret instructions and output spatial coordinates in the standard $[x_{min}, y_{min}, x_{max}, y_{max}]$ format. This phase ensures that the model masters basic task logic and syntax across mobile, and PC platforms before entering the reinforcement stage.

\noindent\textbf{(2) Reinforcement Learning for Spatial Precision:} 
Building upon the SFT baseline, we employ reinforcement fine-tuning using the GRPO framework. GRPO enhances training stability by estimating baselines through relative rewards within groups, significantly reducing the computational overhead typically associated with a separate critic model. Specifically, for each training prompt $q$, GRPO samples a group of $G$ rollouts $\{o_1, o_2, ..., o_G\}$ from the old policy $\pi_{\theta_{old}}$. The advantage $\hat{A}_i$ for each rollout is computed by normalizing the rewards within the group:
\begin{equation}
    \hat{A}_i = \frac{r_i - \text{mean}(\{r_1, r_2, \dots, r_G\})}{\text{std}(\{r_1, r_2, \dots, r_G\})}
\end{equation}
The policy is then optimized by maximizing the following objective function:
\begin{equation}
    \mathcal{J}_{GRPO}(\pi_{\theta}) = \mathbb{E}_{q \sim \mathcal{Q}, \{o_i\}_{i=1}^G \sim \pi_{\theta_{old}}} \left[ \frac{1}{G} \sum_{i=1}^G \left( \frac{1}{|o_i|} \sum_{t=1}^{|o_i|} \mathcal{L}_{clip} (\theta) - \beta D_{KL}(\pi_{\theta} || \pi_{ref}) \right) \right]
\end{equation}
where $\mathcal{L}_{clip}(\theta)$ represents the surrogate objective with a clipping mechanism to prevent excessive policy updates, and the KL divergence term with coefficient $\beta$ constrains the policy from diverging from the reference model $\pi_{ref}$. 
For grounding task, we select a classic dual-component reward function to calibrate the model's spatial perception~\cite{gu2025ui}:

    \hspace*{1em}  \textbf{1). Format Reward ($R_{fmt}$):} A binary reward that validates whether the predicted string conforms to the predefined syntax, ensuring the model outputs executable and parsable responses.
    
    \hspace*{1em}  \textbf{2). Inside-of-Bounding-Box Reward ($R_{pos}$):} This reward targets localization accuracy by incentivizing the model to predict a center point $(x, y)$ that falls strictly within the ground-truth interactive region $[x_{min}, y_{min}, x_{max}, y_{max}]$. The reward is formulated as follows:
    \begin{equation}
        R_{pos} = \begin{cases} 1 & \text{if } x_1 \le x \le x_2 \text{ and } y_1 \le y \le y_2 \\ 0 & \text{otherwise} \end{cases} 
    \end{equation}

   \hspace*{1em} \textbf{3). Total Reward Balancing:} 
To synchronize structural correctness with action precision, the final action-wise reward is computed as a weighted combination:
\begin{equation}
    R = R_{fmt} \cdot w_1 + R_{pos} \cdot w_2
\end{equation}
By carefully balancing the weights $w_1$ and $w_2$, we prevent potential reward conflicts where the model might sacrifice format for precision or vice-versa, ultimately leading to a more robust and coherent grounding policy.

\subsection{OmegaUse: Advanced Planning and Navigation}

In this section, we detail the design and training of OmegaUse’s navigation model, thereby operationalizing our high-quality data construction and decoupled training paradigm. We first present a hierarchical navigation data pipeline that integrates three complementary sources: (1) rigorously curated open-source datasets, (2) automatically synthesized trajectories via bottom-up autonomous exploration and top-down taxonomy-guided generation, and (3) high-fidelity cross-terminal expert demonstrations. We then describe a two-stage optimization strategy, consisting of SFT to establish foundational interaction syntax and task logic, followed by GRPO with specialized reward designs to refine spatial grounding and sequential decision-making.

\subsubsection{Unified Action Space}

To ensure consistent navigation across diverse platforms, we propose a unified action space that standardizes interaction primitives across mobile, desktop, and web platforms. This design organizes agent operations hierarchically, with a core set of shared actions for universal GUI interaction and platform-specific extensions tailored to the unique affordances of each terminal.

As detailed in Table~\ref{tab:unified_action_space}, the shared primitives establish a cross-platform baseline (e.g., click, drag, and type), while specialized actions address terminal-unique requirements—such as desktop hotkeys or mobile system gestures. By harmonizing these disparate operational schemas into a single cohesive space, the model achieves robust cross-terminal generalization. When synchronized with our hierarchical task taxonomy, this architecture enables the agent to execute complex trajectories with unified logical reasoning regardless of the underlying digital ecosystem.

\begin{table}[!t]
\centering
\caption{Unified Action Space Of OmegaUse Across Different Platforms.}
\label{tab:unified_action_space}
\small
\begin{tabularx}{\textwidth}{l p{5.5cm} X}
\toprule
\textbf{Platform} & \textbf{Action Schema} & \textbf{Functional Definition} \\ 
\midrule
\rowcolor[gray]{0.95}
& \texttt{Click(box=(x, y))} & Performs a single-tap or left-click at the given coordinates. \\
\rowcolor[gray]{0.95}
& \texttt{Drag(start, end)} & Executes a drag-and-drop sequence from start point (x1, y1) to end point (x2, y2). \\
\rowcolor[gray]{0.95}
& \texttt{Scroll(start, end, dir)} & Scrolls from (x1, y1) to (x2, y2) in the
given direction. \\
\rowcolor[gray]{0.95}
& \texttt{Type(content=`')} & Injects the specified text string into the active input focus. \\
\rowcolor[gray]{0.95}
& \texttt{Wait()} & Suspends execution to allow for UI state synchronization. \\
\rowcolor[gray]{0.95}
\multirow{-6}{*}{\textbf{Shared}} & \texttt{Finished(content=`')} & Terminates the task and returns the final result. \\
\midrule
& \texttt{Hotkey(key=[`', \dots])} & Simulates hardware keyboard combinations. \\
& \texttt{LeftDouble(box=(x, y))} & Executes a double-click at (x, y). \\
\multirow{-3}{*}{\textbf{Desktop}} & \texttt{RightSingle(box=(x, y))} &  Executes a right-click at (x, y). \\
\midrule
\rowcolor[gray]{0.95}
& \texttt{Hover(box=(x, y))} & Moves the mouse cursor to a specific point. \\
\rowcolor[gray]{0.95}
\multirow{-2}{*}{\textbf{Web}} & \texttt{BrowserStop()} & Interrupts the current page loading process. \\
\midrule
& \texttt{LongPress(box=(x, y))} & Long presses at (x, y). \\
& \texttt{PressBack()} & Navigates to the previous screen. \\
& \texttt{PressHome()} & Returns the device to the primary home screen. \\
\multirow{-4}{*}{\textbf{Mobile}} & \texttt{PressEnter()} & Presses the “enter” key. \\
\bottomrule
\end{tabularx}
\end{table}

\subsubsection{Hierarchical Navigation Data Pipeline}
To bridge the gap between low-level visual perception and high-level logical planning, we construct a large-scale, multi-platform navigation dataset using a hierarchical three-pronged approach: (1) rigorous curation of open-source data, (2) automated trajectory synthesis in virtual sandboxes, and (3) high-fidelity expert demonstrations across multiple terminals.

\noindent\textbf{(1) Open-source Data Curation and Auditing:} 
We leverage the AGUVIS~\cite{xu2024aguvis} stage-2 collection to construct our foundational interaction dataset, which aggregates a diverse array of GUI execution trajectories from both mobile and web terminals, such as AITW~\cite{rawles2023androidinthewild} and Mind2Web~\cite{mind2web_2023}. However, these open-source datasets frequently suffer from significant noise, including misaligned coordinates and fragmented action chains, which can adversely impact model performance if utilized directly. To mitigate these issues~\cite{gu2025ui}, we implement a two-stage quality control pipeline:

Initially, we apply rule-based filtering to eliminate obvious noise and uninformative samples. This involves: (i) enforcing a minimum trajectory length threshold (e.g., $> 3$ steps) to ensure the presence of sufficient learning signals; and (ii) detecting and discarding trajectories characterized by redundant or repetitive action patterns, which typically indicate agent stalling or unproductive exploration.

Subsequently, we employ MLLMs as a high-level trajectory auditor to perform task-completion verification. For each candidate trajectory, the auditor is provided with the specific user goal and the complete execution trace, which includes step-wise action descriptions paired with their corresponding UI screenshots. By jointly analyzing the linguistic intent of the actions and the visual state transitions, the model judges whether the sequence of operations successfully fulfills the original task. Trajectories identified as incomplete or logically inconsistent are strictly filtered out.

\noindent \textbf{(2) Automated Trajectory Synthesis:} To expand the diversity and robustness of our navigation dataset, we implement an automated synthesis framework within simulation environments. We utilize two complementary strategies to balance dataset coverage and task complexity: an Exploration-driven (Bottom-up) approach for autonomous UI discovery, and a Taxonomy-guided (Top-down) approach for generating sophisticated tasks based on expert knowledge.

\begin{figure}[t]
    \centering
    \includegraphics[width=\textwidth]{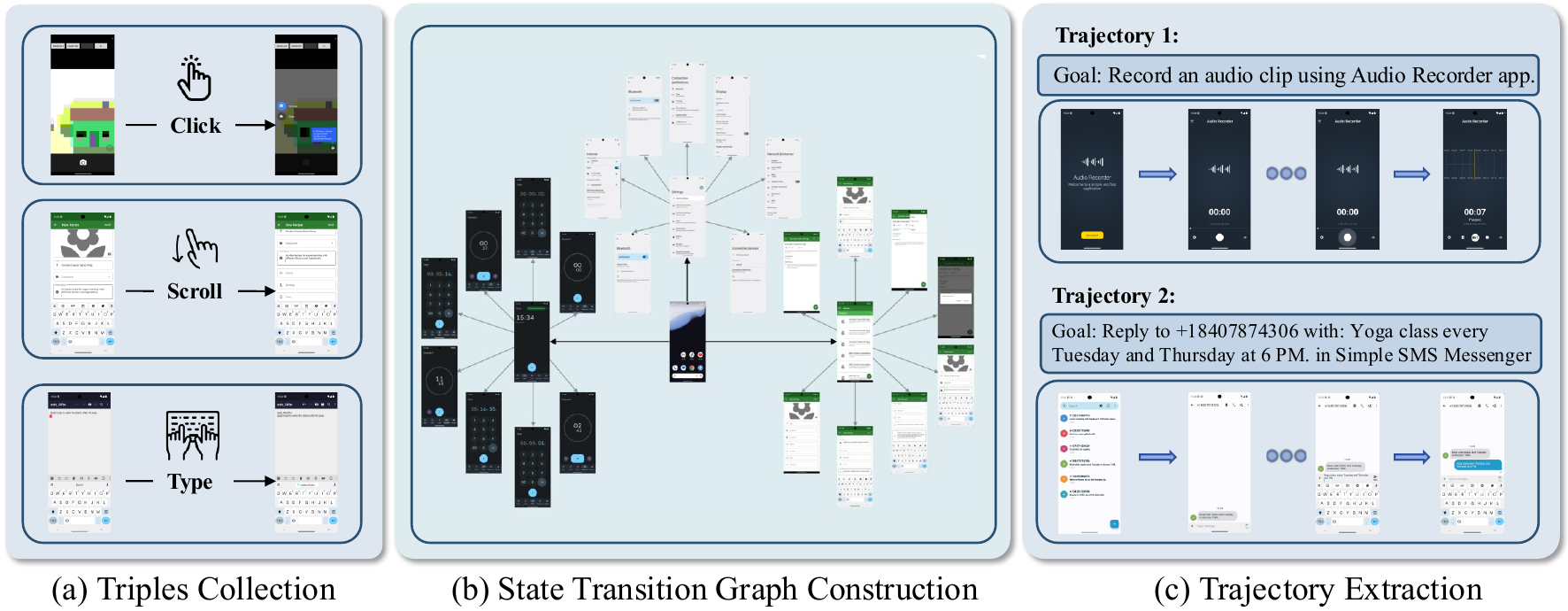} 
    \caption{Overview of the Exploration-driven (Bottom-up) data construction pipeline. (a) Triples Collection: Gathering raw interaction primitives $< pre\_state, action, post\_state >$ through autonomous application exploration. (b) State Transition Graph Construction: Organizing interaction traces into a structured graph with MLLM-based semantic clustering to merge redundant UI states. (c) Trajectory Extraction: Sampling diverse execution paths while enriching them with natural language task goals and step-wise action interpretations.}
    \label{fig:auto_data_instruction}
\end{figure}

\hspace*{1em} \textbf{ Exploration-driven Synthesis (Bottom-up):} To overcome the critical challenges of designing realistic task goals and obtaining diverse execution paths, we implement a systematic bottom-up data construction pipeline as illustrated in Figure~\ref{fig:auto_data_instruction}.  This approach automates high-quality data generation through a four-stage process: interaction exploration, state aggregation, trajectory extraction, and semantic enrichment. Taking the mobile environment as an example, we employ a Depth-First Search (DFS) strategy to explore individual applications within a simulator~\cite{li2017droidbot}. By interacting with UI elements parsed from the Accessibility Tree, the agent collects raw interaction samples in the form of triples: $<pre\_state, action, post\_state>$. Each state is captured as a screenshot, while actions describe specific user behaviors such as clicking or text input. A unique identification and hashing mechanism based on UI structures and action encoding is utilized to avoid redundant exploration of historical states.

To organize these exploration traces into a structured framework, we construct a state transition graph, where each node represents a unique UI state and each directed edge denotes a specific action leading from one state to another. Recognizing the potential for structural redundancy in raw graphs, we introduce a MLLMs-based state clustering and compression mechanism. MLLMs are utilized to perform semantic understanding of screenshots, enabling the system to judge whether multiple nodes belong to the same functional page, such as ``Settings" pages with minor visual variations. These redundant nodes are merged into virtual nodes to reduce the graph scale and significantly improve subsequent computational efficiency.

Based on the refined state transition graph, we perform trajectory extraction by enumerating multiple reachable paths from the initial state. To ensure the logical coherence of the generated data, we implement a cycle-avoidance strategy that maintains a visit set for each path, skipping branches that would lead to unproductive UI loops, such as ``Settings $\rightarrow$ Back $\rightarrow$ Settings". Following trajectory extraction, we utilize MLLMs for semantic enrichment at two levels: action interpretation and task goal generation. Each triple is translated into a natural language description, such as ``Click the `Settings' button in the top right corner", while the entire action sequence is abstracted into a coherent task objective, like ``Modify notification permissions in the settings menu". This mapping from execution trajectories to high-level linguistic goals provides the core supervision signal required to train the agent for robust instruction-to-action generation.

\hspace*{1em}\textbf{Taxonomy-guided Generation (Top-down):} We propose a taxonomy-guided generation framework and apply it across desktop and mobile environments to ensure comprehensive coverage of diverse real-world interaction behaviors. For each kind of environment, we design a specialized hierarchical task taxonomy grounded in its unique ecosystem and typical usage patterns. Guided by these taxonomies, task descriptions are generated and subsequently executed by a high-capability expert model within our unified simulation environments. The agent performs self-assessment of execution correctness based on real-time environmental feedback, and trajectories with successful outcomes are recorded as candidate samples. As a representative instance, Table~\ref{tab:task_taxonomy_computer} shows a hierarchical taxonomy developed for typical daily GUI usage patterns in a desktop environment. 
    
To ensure annotation reliability, we build a human-in-the-loop verification platform in which human annotators verify cases that the model has labeled as successful. This design reduces noise, prevents the accumulation of errors from automatic labeling, and improves the overall robustness of the dataset. In addition, we collect two types of failure cases: (i) those that LLM itself judges as unsuccessful and (ii) those it incorrectly judges as successful. These model-generated failure cases are then handed over to human annotators for careful relabeling, resulting in a curated diagnostic failure subset of challenging GUI tasks that even state-of-the-art closed-source models such as LLM fail to solve.

 \textbf{(3) Cross-Terminal Expert Demonstrations:} 
To establish a high-quality data for the navigation model, we also implement an expert demonstration pipeline for desktop and mobile environments.

Based on the hierarchical taxonomies established for each terminal, we utilize LLMs to synthesize task instructions through divergent reasoning. To ensure task depth and challenge, we enforce a strict complexity constraint, requiring each instruction to involve at least five operational steps. Before the annotation phase, human experts manually vet these prompts for logical validity and environmental feasibility, refining or discarding any substandard entries to ensure a high-quality instruction pool.

The verified instructions are subsequently distributed to our proprietary annotation platform, where professional annotators perform step-by-step executions within simulation environments until task completion. This process ensures that the resulting trajectories capture precise state transitions and aligned action sequences.

To guarantee maximum data reliability, we implement a two-tier quality assurance protocol where each completed trajectory must undergo independent audit by two inspectors. Only samples passing both goal alignment and logical consistency checks are finally retained, resulting in a high-fidelity expert demonstration dataset that serves as a robust foundation for model training and benchmarking.

\begin{table}[t]
    \centering
    \caption{Hierarchical task taxonomy for desktop GUI Navigation. This taxonomy guides the top-down generation process to ensure a diverse coverage of real-world user scenarios.}
    \label{tab:task_taxonomy_computer}
    \small 
    \begin{tabularx}{\textwidth}{l X}
        \toprule
        \textbf{Domain} & \textbf{Core Functionalities and Sub-scenarios} \\
        \midrule
        \rowcolor[gray]{0.95} Desktop Office & Document Editing, Spreadsheet Processing, Presentation Creation, PDF Workflows, Collaboration \& Sharing. \\
        \addlinespace
        Browser \& Web & Tab Management, Privacy \& Security, Browser Extensions, Account Sync, Developer Tools. \\
        \addlinespace
        \rowcolor[gray]{0.95} Communication & Instant Messaging, Meetings \& Remote Collaboration, Email, Calendar Integration. \\
        \addlinespace
        File Management & Search \& Indexing, Compression, Archive Management, Storage Sync, External Media Operations. \\
        \addlinespace
        \rowcolor[gray]{0.95} System Operations & Display \& Device Settings, Network Connectivity, Power \& Updates, Software Management, Notifications \& Focus. \\
        \addlinespace
        Media \& Ent. & Image Editing, Media Playback, Content Library Management. \\
        \addlinespace
        \rowcolor[gray]{0.95} DevOps \& Tech & Development Environments, Version Control, System Technical Operations, Deployment. \\
        \addlinespace
        Productivity Tools & Screen Capture, Notes \& Tasks, Calculator, Time Management, Desktop Enhancements. \\
        \addlinespace
        \rowcolor[gray]{0.95} Security \& Privacy & Account Access Security, System Protection, Encryption, Privacy Shielding. \\
        \bottomrule
    \end{tabularx}
\end{table}

\subsubsection{Two-Stage Navigation Training}

To develop a robust GUI navigation agent capable of complex multi-step planning, we implement a two-stage optimization paradigm. This strategy leverages a massive initial corpus for general behavioral alignment followed by high-precision reinforcement learning on expert-verified data.

\noindent \textbf{(1) Policy Initialization (SFT):} 
The navigation model is first trained using SFT on a diverse dataset of approximately $260K$ instances. This corpus consists of aggregated open-source navigation traces and automatically synthesized trajectories. This stage focuses on teaching the model the fundamental mapping between linguistic goals and cross-platform action sequences, establishing a stable starting policy that adheres to the unified action space.


At each step $t$, the agent receives the multimodal input $X_t = \{I, V_t, H_t\}$, where $I$ is the task instruction, $V_t$ is the current screenshot, and $H_t$ represents the historical reasoning traces. The agent's response is structured as a triplet $Y_t = (O_t, T_t, A_t)$: the observation $O_t$ semantically describes the UI state; the thought $T_t$ performs goal-oriented reasoning based on $I$; and the action $A_t$ provides the executable code snippet conforming to our unified action space. This $O \rightarrow T \rightarrow A$ pipeline ensures each action is grounded in explicit perception and logical planning.

\noindent \textbf{(2) Reinforcement Learning for Decision Robustness:} 
Based on the SFT baseline, we also employ reinforcement learning using the GRPO framework. To provide the fine-grained feedback, we design a multi-dimensional reward function that assesses both structural integrity and operational logic:
    
\hspace*{1em} \textbf{1). Format Reward ($R_{fmt}$):} This reward validates whether the output strictly conforms to the required template, ensuring reasoning and actions are correctly enclosed within structured tags.
        
\hspace*{1em} \textbf{2). Action-wise Reward ($R_{act}$):} This component evaluates the execution logic and is further decomposed into:
        \begin{itemize}
            \item \textbf{Type Accuracy ($R_{type}$):} A binary reward for matching the correct action primitive (e.g., \texttt{Click} vs. \texttt{Scroll}).
            \item \textbf{Coordinate Precision ($R_{coord}$):} For spatial actions, we apply a stepwise reward based on the distance between the predicted and ground-truth coordinates:
            \begin{equation}
                R_{coord} = \begin{cases} 
                1.0 &\text{if } \Delta x, \Delta y < \theta_1, \\ 
                0.5 & \text{if } \theta_1 \le \Delta x, \Delta y < \theta_2, \\ 
                0 & \text{otherwise.} 
                \end{cases}
            \end{equation}
        where $\Delta x = |x_{pred} - x_{gt}|$ and $\Delta y = |y_{pred} - y_{gt}|$ denote the absolute differences between the predicted coordinates and the ground truth along the $x$ and $y$ axes, respectively. The parameters $\theta_1$ and $\theta_2$ serve as predefined distance thresholds that determine the precision of the agent's spatial grounding during coordinate-based actions such as \texttt{Click} or \texttt{LeftDouble}.

For the \texttt{Drag} action, the reward $R_{\text{drag}}$ is calculated based on the coordinate deviations of both the start and end points:
\begin{equation}
    R_{{drag}} = 
    \begin{cases} 
    1.0 & \text{if } \max(\Delta x_1, \Delta y_1, \Delta x_2, \Delta y_2) \leq \alpha_1 \\
    0.5 & \text{if } \alpha_1 < \max(\Delta x_1, \Delta y_1, \Delta x_2, \Delta y_2) \leq \alpha_2 \\
    0 & \text{otherwise}
    \end{cases}
\end{equation}
where $\Delta x_i = |x_{i, \text{pred}} - x_{i, \text{gt}}|$ and $\Delta y_i = |y_{i, \text{pred}} - y_{i, \text{gt}}|$ represent the absolute errors for the start ($i=1$) and end ($i=2$) coordinates.

For the \texttt{Scroll} action, the reward $R_{\text{scroll}}$ incorporates both spatial precision and directional accuracy:
\begin{equation}
    R_{{scroll}} = 
    \begin{cases} 
    1.0 & \text{if } \max(\Delta x_1, \Delta y_1, \Delta x_2, \Delta y_2) \leq \beta_1 \text{, and } \text{dir}_{\text{pred}} = \text{dir}_{\text{gt}} \\
    0.5 & \text{if } \beta_1 < \max(\Delta x_1, \Delta y_1, \Delta x_2, \Delta y_2) \leq \beta_2 \text{ ,and } \text{dir}_{\text{pred}} = \text{dir}_{\text{gt}} \\
    0 & \text{otherwise}
    \end{cases}
\end{equation}
where $\text{dir}_{\text{pred}}$ and $\text{dir}_{\text{gt}}$ denote the predicted and ground-truth scroll directions. This formulation ensures the agent's scrolling behavior is both spatially grounded and semantically correct.

\item \textbf{Content Fidelity ($R_{content}$):} For typing tasks, the reward is determined by the  token-level F1-score of the predicted string $S_1$ relative to the ground-truth target $S_2$

\begin{equation}
    R_{content} = \begin{cases} 
    1.0 &\text{if F1-score} \geq 0.5 , \\ 
    0 & \text{otherwise.} 
    \end{cases}
\end{equation}

For the \texttt{Hotkey} action, the reward $R_{\text{hotkey}}$ is defined by a binary matching criterion, requiring the predicted key combination to be identical to the ground truth:

\begin{equation}
    R_{hotkey} = \begin{cases} 
    1.0 &\text{if } K_{\text{pred}} = K_{\text{gt}} , \\ 
    0 & \text{otherwise.} 
    \end{cases}
\end{equation}
where $K_{\text{pred}}$ and $K_{\text{gt}}$ represent the predicted and ground-truth hotkey parameter sets (e.g., {[`\texttt{ctrl}', `\texttt{c}']}). Given that hotkey operations are sensitive to exact key combinations, this strict matching ensures the agent executes the precise system-level command intended.

\end{itemize}

\hspace*{1em} \textbf{3). Total Reward Balancing:} 
    The final reward for each step is a weighted sum that balances structural consistency with action accuracy:
    \begin{equation}
        R = R_{fmt} \cdot w_{3} + R_{act} \cdot w_{4}
    \end{equation}
    where $w_3$ and $w_4$ are hyper-parameters tuned to prevent the model from sacrificing action precision for format compliance or vice-versa.


\section{Offline Benchmarks for Real-World GUI Navigation}

To facilitate evaluation of agent performance in realistic digital environments, we introduce \textbf{OS-Nav}, a specialized offline benchmark comprising two sub-benchmarks across different operating systems: \textbf{ChiM-Nav}, focusing on Chinese Android mobile systems, and \textbf{Ubu-Nav}, targeting routine desktop interactions on Ubuntu. The benchmark is open-sourced, and can be publicly accessed
\footnote{https://huggingface.co/datasets/baidu-frontier-research/OS-Nav}.

To ensure the reliability of state transitions and the transparency of agent logic, both benchmarks were developed using a rigorous human-AI collaborative pipeline. We curated expert-labeled execution traces to ensure all tasks reflect authentic user behavior. For every step, we utilized MLLMs to synthesize intermediate CoT descriptions, providing a semantic bridge between linguistic goals and raw actions. Every trajectory, including the AI-generated reasoning, underwent final refinement by human experts to ensure the ``gold" labels are logically sound and environment-feasible.

\subsection{ChiM-Nav: Chinese Mobile Navigation Benchmark}

The ChiM-Nav benchmark assesses an agent's ability to navigate popular applications within the Chinese mobile ecosystem. This suite comprises \textbf{142} trajectories across \textbf{69} distinct applications, totaling \textbf{991} operational steps. With an average trajectory length of \textbf{6.98} steps, the benchmark emphasizes daily usage scenarios and evaluates the agent's robustness against the unique UI layouts and multi-step workflows characteristic of Chinese digital platforms.

\subsection{Ubu-Nav: General Desktop Navigation Benchmark}

The Ubu-Nav benchmark consists of \textbf{101} trajectories with a total of \textbf{641} steps, targeting agent performance in Ubuntu environments. Trajectories in this benchmark range from \textbf{2 to 11} steps, with an average length of \textbf{6.35} steps per task. It covers extensive routine desktop operations and typical system interactions, focusing on the multi-step reasoning required for common PC tasks.

\section{Experiments}


In this section, we evaluate OmegaUse on a set of grounding and navigation benchmarks across mobile and desktop platforms. Our experiments validate the contributions of our high-quality data construction pipeline, the decoupled training strategy, and cross-terminal generalization on OS-Nav.

\begin{table}[t]
\centering
\caption{Performance comparison on ScreenSpot-V2 dataset. The Avg. column represents the overall success rate across all categories.}\label{tab:screenspot_v2}
\resizebox{0.9\textwidth}{!}{
\begin{tabular}{l cc cc cc c}
\toprule
\multirow{2}{*}{\textbf{Models}}&\multicolumn{2}{c}{\textbf{Mobile}} & \multicolumn{2}{c}{\textbf{Desktop}} & \multicolumn{2}{c}{\textbf{Web}} & \multirow{2}{*}{\textbf{Avg}} \\
\cmidrule(lr){2-3} \cmidrule(lr){4-5} \cmidrule(lr){6-7}
 & Text & Icon/Widget & Text & Icon/Widget & Text & Icon/Widget & \\
\midrule
\rowcolor[gray]{0.95}\multicolumn{8}{l}{\textit{Closed-source Models}} \\
GPT-4o \cite{islam2025gpt} & 26.6 & 24.2 & 24.2 & 19.3 & 12.8 & 11.8 & 20.1 \\
UI-TARS-1.5 \cite{ui-tars-15-seed} & - & - & - & - & - & - & 94.2 \\
Seed1.5-VL \cite{guo2025seed1} & - & - & - & - & - & - & \underline{95.2} \\
\midrule
\rowcolor[gray]{0.95}\multicolumn{8}{l}{\textit{GUI-specific Models (SFT)}} \\
SeeClick-9.6B \cite{cheng2024seeclick} & 78.4 & 50.7 & 70.1 & 29.3 & 55.2 & 32.5 & 55.1 \\
ShowUI-2B \cite{lin2025showui} & 92.1 & 75.4 & 78.9 & 78.9 & 84.2 & 61.1 & 77.3 \\
UGround-7B \cite{qian2025uground} & 75.1 & 84.5 & 85.1 & 61.4 & 84.6 & 71.9 & 76.3 \\
OS-Atlas-7B \cite{wu2024osatlas} & 95.2 & 75.8 & 90.7 & 63.6 & 90.6 & 77.3 & 84.1 \\
Aguvis-7B \cite{xu2024aguvis} & 89.3 & 68.7 & 80.6 & 67.9 & 89.3 & 70.0 & 80.5 \\
UI-TARS-7B \cite{ui_tars_2025} & 96.9 & 89.1 & 95.4 & 85.0 & 93.6 & 85.2 & 91.6 \\
UI-TARS-72B \cite{ui_tars_2025} & 94.8 & 86.3 & 91.2 & 87.9 & 91.5 & 87.7 & 90.3 \\
JEDI-7B \cite{xie2025scaling} & 96.9 & 87.2 & 95.9 & 87.9 & 94.4 & 84.2 & 91.7 \\
GUI-Actor-7B \cite{wu2025gui} & 97.6 & 88.2 & 96.9 & 85.7 & 93.2 & 86.7 & 92.1 \\
OpenCUA-7B \cite{wang2025opencua} & - & - & - & - & - & - & 92.3 \\
OpenCUA-32B \cite{wang2025opencua} & - & - & - & - & - & - & 93.4 \\
\midrule
\rowcolor[gray]{0.95}\multicolumn{8}{l}{\textit{GUI-specific Models (RL)}} \\
UI-R1-E-3B \cite{lu2025ui} & 98.2 & 83.9 & 94.8 & 75.0 & 93.2 & 83.7 & 89.5 \\
SE-GUI-7B \cite{yuan2025enhancing} & - & - & - & - & - & - & 90.3 \\
LPO \cite{tang2025lpo} & 97.9 & 82.9 & 95.9 & 86.4 & 95.6 & 84.2 & 90.5 \\
GUI-G$^2$-7B \cite{tang2025guig2} & - & - & - & - & - & - & 93.3 \\
Phi-Ground-7B-16C-DPO \cite{zhang2025phi} & 96.5 & 62.0 & 90.2 & 76.4 & 93.6 & 75.9 & 83.8 \\
GTA1-7B† \cite{yang2025gta1} & 99.0 & 88.6 & 94.9 & 89.3 & 92.3 & 86.7 & 92.4 \\
GTA1-72B \cite{yang2025gta1} & \underline{99.3} & 92.4 & \underline{97.4} & 89.3 & 95.3 & 91.4 & 94.8 \\
UI-Venus-Ground-7B \cite{gu2025ui} & 99.0 & 90.0 & 97.0 & \underline{90.7} & \underline{96.2} & 88.7 & 94.1 \\
UI-Venus-Ground-72B \cite{gu2025ui} & \textbf{99.7} & \underline{93.8} & 95.9 & 90.0 & \underline{96.2} & \underline{92.6} &  \underline{95.3} \\
\midrule
\textbf{OmegaUse-G} & \underline{99.3} & \textbf{94.3} & \textbf{99.0} & \textbf{96.4} & \textbf{97.4} & \textbf{94.0} & \textbf{96.3} \\
\bottomrule
\end{tabular}
}
\end{table}

\subsection{Experimental Setup}
\subsubsection{Model Configurations}
We employed a 30B-A3B VL model as the backbone of OmegaUse. In the SFT phase, we fine-tuned the model for one epoch using a learning rate of $1e^{-5}$, a global batch size of $32$, and a temperature of $1.0$. For the subsequent Grounding and Navigation RL phase, we trained for one epoch with a learning rate of $5e^{-5}$, a global batch size of $64$, and a temperature of $1.0$. Specifically for RL, we utilized 8 sampled responses per instruction and set the KL penalty coefficient $\beta$ to $0.04$. Across both phases, we maintained an MoE auxiliary loss coefficient of $1e^{-6}$ and a maximum image token limit of $16,384$.

\subsection{Evaluation of GUI Grounding}

We evaluate the grounding performance of our OmegaUse model across two major benchmarks: \textbf{ScreenSpot-V2} and \textbf{ScreenSpot-Pro}. These benchmarks test the model's ability to associate natural language instructions with diverse UI elements across mobile, web, and desktop platforms.

\begin{table}[t]
\centering
\caption{Performance comparison of different agent models on ScreenSpot-Pro. The Avg. column represents the overall success rate across all categories.}
\label{tab:performance_comparison}
\resizebox{\textwidth}{!}{
\begin{tabular}{l cccccccccccc c}
\toprule
\multirow{2}{*}{\textbf{Model}} & \multicolumn{2}{c}{\textbf{CAD}} & \multicolumn{2}{c}{\textbf{Dev}} & \multicolumn{2}{c}{\textbf{Creative}} & \multicolumn{2}{c}{\textbf{Scientific}} & \multicolumn{2}{c}{\textbf{Office}} & \multicolumn{2}{c}{\textbf{OS}} & \multirow{2}{*}{\textbf{Avg.}} \\
\cmidrule(lr){2-3} \cmidrule(lr){4-5} \cmidrule(lr){6-7} \cmidrule(lr){8-9} \cmidrule(lr){10-11} \cmidrule(lr){12-13}
 & Text & Icon & Text & Icon & Text & Icon & Text & Icon & Text & Icon & Text & Icon & \\
\midrule
\rowcolor[gray]{0.95}\multicolumn{14}{l}{\textit{Closed-source Models}} \\
GPT-4o \cite{islam2025gpt} & 2.0 & 0.0 & 1.3 & 0.0 & 1.0 & 0.0 & 2.1 & 0.0 & 1.1 & 0.0 & 0.0 & 0.0 & 0.8 \\
Claude Computer Use \cite{anthropic2024computeruse} & 14.5 & 3.7 & 22.0 & 3.9 & 25.9 & 3.4 & 33.9 & 15.8 & 30.1 & 16.3 & 11.0 & 4.5 & 17.1 \\
UI-TARS-1.5 \cite{ui-tars-15-seed} & - & - & - & - & - & - & - & - & - & - & - & - & \underline{61.6} \\
Seed1.5-VL \cite{guo2025seed1} & - & - & - & - & - & - & - & - & - & - & - & - & 60.9 \\
\midrule
\rowcolor[gray]{0.95}\multicolumn{14}{l}{\textit{GUI-specific Models (SFT)}} \\
SeeClick-9.6B \cite{cheng2024seeclick} & 2.5 & 0.0 & 0.6 & 0.0 & 1.0 & 0.0 & 3.5 & 0.0 & 1.1 & 0.0 & 2.8 & 0.0 & 1.1 \\
FOCUS-2B \cite{tang2025think} & 7.6 & 3.1 & 22.8 & 1.7 & 23.7 & 1.7 & 25.0 & 7.1 & 23.2 & 7.7 & 17.8 & 2.5 & 13.3 \\
CogAgent-18B \cite{hong2024cogagent} & 7.1 & 3.1 & 14.9 & 0.7 & 9.6 & 0.0 & 22.2 & 1.8 & 13.0 & 0.0 & 5.6 & 0.0 & 7.7 \\
Aria-UI \cite{yang2025aria} & 7.6 & 1.6 & 16.2 & 0.0 & 23.7 & 2.1 & 27.1 & 6.4 & 20.3 & 1.9 & 4.7 & 0.0 & 11.3 \\
OS-Atlas-7B \cite{wu2024osatlas} & 12.2 & 4.7 & 33.1 & 1.4 & 28.8 & 2.8 & 37.5 & 7.3 & 33.9 & 5.7 & 27.1 & 4.5 & 18.9 \\
ShowUI-2B \cite{lin2025showui} & 2.5 & 0.0 & 16.9 & 1.4 & 9.1 & 0.0 & 13.2 & 7.3 & 15.3 & 7.5 & 10.3 & 2.2 & 7.7 \\
UGround-7B \cite{qian2025uground} & 14.2 & 1.6 & 26.6 & 2.1 & 27.3 & 2.8 & 31.9 & 2.7 & 31.6 & 11.3 & 17.8 & 0.0 & 16.5 \\
UGround-V1-7B \cite{qian2025uground} & 15.8 & 1.2 & 51.9 & 2.8 & 47.5 & 9.7 & 57.6 & 14.5 & 60.5 & 13.2 & 38.3 & 7.9 & 31.1 \\
UI-TARS-7B \cite{ui_tars_2025} & 20.8 & 9.4 & 58.4 & 12.4 & 50.0 & 9.1 & 63.9 & 31.8 & 63.3 & 20.8 & 30.8 & 16.9 & 35.7 \\
UI-TARS-72B \cite{ui_tars_2025} & 18.8 & 12.5 & 62.9 & 17.2 & 57.1 & 15.4 & 64.6 & 20.9 & 63.3 & 26.4 & 42.1 & 15.7 & 38.1 \\
JEDi-7B \cite{xie2025scaling} & 38.0 & 14.1 & 42.9 & 11.0 & 50.0 & 11.9 & 72.9 & 25.5 & 75.1 & 47.2 & 33.6 & 16.9 & 39.5 \\
GUI-Actor-7B \cite{wu2025gui} & - & - & - & - & - & - & - & - & - & - & - & - & 44.6 \\
OpenCUA-7B \cite{wang2025opencua} & - & - & - & - & - & - & - & - & - & - & - & - & 50.0 \\
OpenCUA-32B \cite{wang2025opencua} & - & - & - & - & - & - & - & - & - & - & - & - & 55.3 \\
\midrule
\rowcolor[gray]{0.95}\multicolumn{14}{l}{\textit{GUI-specific Models (RL)}} \\
UI-R1-E-3B \cite{lu2025ui} & 37.1 & 12.5 & 46.1 & 6.9 & 41.9 & 4.2 & 56.9 & 21.8 & 65.0 & 26.4 & 32.7 & 10.1 & 33.5 \\
UI-R1-7B \cite{lu2025ui} & 23.9 & 6.3 & 49.4 & 4.8 & 38.9 & 8.4 & 55.6 & 11.8 & 58.7 & 26.4 & 42.1 & 16.9 & - \\
InfiGUI-R1-3B \cite{liu2025infigui} & 33.0 & 14.1 & 51.3 & 12.4 & 44.9 & 7.0 & 58.3 & 20.0 & 65.5 & 28.3 & 43.9 & 12.4 & 35.7 \\
GUI-G1-3B \cite{zhou2025guig1} & 39.6 & 9.4 & 50.7 & 10.3 & 36.6 & 11.9 & 61.8 & 30.0 & 67.2 & 32.1 & 23.5 & 10.6 & 37.1 \\
SE-GUI-7B \cite{yuan2025enhancing} & 51.3 & \textbf{42.2} & 68.2 & 19.3 & 57.6 & 9.1 & 75.0 & 28.2 & 78.5 & 43.4 & 49.5 & 25.8 & 47.3 \\
Phi-Ground-7B-16C-DPO \cite{zhang2025phi} & 26.9 & 17.2 & 70.8 & 16.7 & 56.6 & 13.3 & 58.0 & 29.1 & 76.4 & 44.0 & 55.1 & 25.8 & 43.2 \\
GUI-G$^2$-7B  \cite{tang2025guig2} & 55.8 & 12.5 & 68.8 & 17.2 & 57.1 & 15.4 & 77.1 & 24.5 & 74.0 & 32.7 & 57.9 & 21.3 & 47.5 \\
UI-TARS-1.5-7B \cite{ui-tars-15-seed} & - & - & - & - & - & - & - & - & - & - & - & - & 49.6 \\
GTA1-7B\dag \cite{yang2025gta1} & 53.3 & 17.2 & 66.9 & 20.7 & 62.6 & 18.2 & 76.4 & 31.8 & 82.5 & \underline{50.9} & 48.6 & 25.9 & 50.1 \\
GTA1-72B \cite{yang2025gta1} & 56.9 & 28.1 & \underline{79.9} & \textbf{33.1} & \textbf{73.2} & 20.3 & \underline{81.9} & \underline{38.2} & \textbf{85.3} & 49.1 & {73.8} & \underline{39.1} & 58.4 \\
UI-Venus-Ground-7B \cite{gu2025ui} & \underline{60.4} & 21.9 & 74.7 & 24.1 & 63.1 & 14.7 & 76.4 & 31.8 & 75.7 & 41.5 & 49.5 & 22.5 & 50.8 \\
UI-Venus-Ground-72B \cite{gu2025ui} & \textbf{66.5} & \underline{29.7} & \textbf{84.4} & \textbf{33.1} & \textbf{73.2} & \textbf{30.8} & \textbf{84.7} & \textbf{42.7} & \underline{83.1} & \textbf{60.4} & \textbf{75.7} & 36.0 & \textbf{61.9} \\
\midrule
\textbf{OmegaUse-G} & 48.73 & 23.44 & 78.57 & \underline{31.72} & \underline{66.67} & \underline{22.38} & 75.69 & 34.55 & 81.36 & 47.17 & \underline{74.77} & \textbf{43.82} & 55.47 \\
\bottomrule
\end{tabular}
}
\end{table}

\textbf{ScreenSpot-V2.} As a fundamental GUI grounding benchmark, ScreenSpot-V2 measures the agent's localization reliability across mobile, web, and desktop interfaces. As shown in Table~\ref{tab:screenspot_v2}, {OmegaUse} achieves an exceptional state-of-the-art average score of \textbf{96.3\%}, establishing a new performance ceiling for this benchmark. It consistently outperforms leading baselines, including UI-Venus-Ground-72B (95.3\%) and Seed1.5-VL (95.2\%). A detailed breakdown reveals that {OmegaUse-G} maintains near-perfect accuracy on text-based elements, particularly in the mobile and desktop segments, where it scores 99.3\% and 99.0\%, respectively. Furthermore, its performance on icon and widget localization remains remarkably high, reaching 96.4\% on desktop and 94.0\% on web platforms, demonstrating robust cross-platform generalization and precise spatial perception.

\textbf{ScreenSpot-Pro.} Compared to standard GUI grounding benchmarks, ScreenSpot-Pro presents a more rigorous evaluation by featuring high-resolution interfaces from professional software, often characterized by intricate and microscopic visual elements. In this challenging setting, as detailed in Table~\ref{tab:performance_comparison}, OmegaUse-G achieves a competitive average score of 55.47\%.

While ultra-large-scale models such as UI-Venus-Ground-72B (61.9\%) and GTA1-72B (58.4\%) maintain a lead in overall performance, \text{OmegaUse} demonstrates specialized strengths in specific domains. Notably, it achieves the highest accuracy in the {OS-Icon} category ({43.82\%}), outperforming all baseline models. Furthermore, it attains runner-up performance in several key metrics, including {74.77\%} in OS-Text, {31.72\%} in Dev-Icon, and {66.67\%} in Creative-Text. These results indicate that despite a smaller parameter scale compared to 72B-class models, \text{OmegaUse-G} exhibits robust precision in professional and system-level GUI environments, particularly in capturing fine-grained icon details and complex text layouts within creative and developer tools.

\subsection{Evaluation of GUI Navigation}
Navigation performance is evaluated on both widely used standard benchmarks and our specialized offline benchmark OS-Nav.

\subsubsection{Standard Benchmark}

We evaluate the multi-step decision-making and planning capabilities of {OmegaUse} across two widely-adopted benchmarks: \textbf{AndroidControl}~\cite{li2024effects} for offline trajectory planning and \textbf{AndroidWorld}~\cite{rawles2024androidworld} for online interaction. These evaluations assess the model's ability to translate high-level user goals into coherent, executable action sequences.

\begin{table}[t]
\centering
\small
\caption{Performance comparison on the \textbf{AndroidControl} offline UI navigation dataset.}
\label{tab:ui_navigation_reduced} 
\resizebox{0.6\textwidth}{!}{
\begin{tabular}{l cc}
\toprule
\textbf{Model} & \textbf{Type Acc. (\%)} & \textbf{Step SR (\%)} \\
\midrule
\rowcolor[gray]{0.95}\multicolumn{3}{l}{\textit{Open-source Models}} \\
SeeClick \cite{cheng2024seeclick}& 82.9 & 59.1 \\
OS-Atlas-7B \cite{wu2024osatlas} & 85.2 & 71.2 \\
Aguvis-7B \cite{xu2024aguvis} & -- & 61.5 \\
Aguvis-72B \cite{xu2024aguvis} & -- & 66.4 \\
OS-Genesis-7B \cite{sun2025osgenesis} & 66.2 & 44.5 \\
UI-TARS-7B \cite{ui_tars_2025} & 83.7 & 72.5 \\
UI-TARS-72B \cite{ui_tars_2025} & 85.2 & 74.7 \\
GUI-R1-7B \cite{luo2025gui} & 71.6 & 51.7 \\
NaviMaster-7B \cite{luo2025navimaster} & 72.9 & 54.0 \\
UI-AGILE-7B \cite{lian2025ui} & 80.1 & 60.6 \\
AgentCPM-GUI \cite{zhang2025agentcpm} & 77.7 & 69.2 \\
UI-Venus-Navi-7B \cite{gu2025ui} & \underline{86.5} & 76.1 \\
UI-Venus-Navi-72B \cite{gu2025ui} & 85.9 & \underline{77.2} \\
\midrule
\rowcolor[gray]{0.95} \textbf{OmegaUse} & \textbf{87.6} & \textbf{79.1} \\
\bottomrule
\end{tabular}
}
\end{table}

\begin{table}[t]
\centering
\caption{Performance comparison on \textbf{AndroidWorld} for end-to-end models.}
\label{tab:android_world} 
\resizebox{0.8\textwidth}{!}{ 
\begin{tabular}{l cccc}
\toprule
\textbf{Models} & \textbf{Planner} & \textbf{A11y Tree} & \textbf{Screenshot} & \textbf{Success Rate} \\
\midrule
\rowcolor[gray]{0.95} \multicolumn{5}{l}{\textit{Closed-source Models}} \\
GPT-4o \cite{islam2025gpt} & $\times$ & $\checkmark$ & $\times$ & 30.6 \\
ScaleTrack \cite{huang2025scaletrack} & $\times$ & $\checkmark$ & $\times$ & 44.0 \\
SeedVL-1.5 \cite{guo2025seed1} & $\times$ & $\checkmark$ & $\checkmark$ & 62.1 \\
UI-TARS-1.5 \cite{ui-tars-15-seed} & $\times$ & $\times$ & $\checkmark$ & 64.2 \\
\midrule
\rowcolor[gray]{0.95}\multicolumn{5}{l}{\textit{Open-source Models}} \\
GUI-Critic-R1-7B \cite{wanyan2025look} & $\times$ & $\checkmark$ & $\checkmark$ & 27.6 \\
Qwen2.5-VL-72B \cite{bai2025qwen25vl} & $\times$ & $\times$ & $\checkmark$ & 35.0 \\
UGround \cite{qian2025uground} & $\checkmark$ & $\times$ & $\checkmark$ & 44.0 \\
Aria-UI \cite{yang2025aria} & $\checkmark$ & $\times$ & $\checkmark$ & 44.8 \\
UI-TARS-72B  \cite{ui_tars_2025} & $\times$ & $\times$ & $\checkmark$ & 46.6 \\
GLM-4.5v \cite{glmvteam2025glm41vthinkingversatilemultimodalreasoning} & $\times$ & $\times$ & $\checkmark$ & 57.0 \\
\addlinespace 
UI-Venus-Navi-7B \cite{gu2025ui} & $\times$ & $\times$ & $\checkmark$ & 49.1 \\
UI-Venus-Navi-72B \cite{gu2025ui} & $\times$ & $\times$ & $\checkmark$ & \textbf{65.9} \\
\midrule
\textbf{OmegaUse} & $\times$ & $\times$ & $\checkmark$ & 55.7 \\
\bottomrule
\end{tabular}
}
\end{table}

\begin{table}[htbp]
\centering
\small
\caption{Performance comparison on the \textbf{ChiM-Nav} offline navigation dataset.}
\label{tab:ChiM-Nav}
\resizebox{0.6\textwidth}{!}{
\begin{tabular}{l cc}
\toprule
\textbf{Model} & \textbf{Type Acc. (\%)} & \textbf{Step SR (\%)} \\
\midrule
\rowcolor[gray]{0.95}\multicolumn{3}{l}{\textit{Open-source Models}} \\
UI-TARS-SFT \cite{ui_tars_2025} & 53.28 & 36.97 \\
UI-TARS-1.5 \cite{ui-tars-15-seed} & 64.12 & 37.24 \\
GUI-R1-7B \cite{luo2025gui} & 63.74 & 34.74 \\
OS-Atlas-7B  \cite{wu2024osatlas} & 59.63 & 38.26 \\
UI-AGILE-7B  \cite{lian2025ui} & 70.2 & 45.96 \\
AgentCPM-GUI \cite{zhang2025agentcpm} & 75.02 & 51.62 \\
Holo2-30b-A3B \cite{hai2025holo2modelfamily} & 73.76 & 60.69 \\
Qwen3-VL-30b-A3B \cite{bai2025qwen3vl} & 78.2 & 65.19 \\
UI-Venus-72B \cite{gu2025ui} & 81.23 & 67.51 \\
Qwen3-VL-32B \cite{bai2025qwen3vl} & 80.83 & 66.39 \\
\midrule
\rowcolor[gray]{0.95} \textbf{OmegaUse} & \textbf{87.78} & \textbf{74.24} \\
\bottomrule
\end{tabular}
}
\end{table}

\begin{table}[htbp]
\centering
\small
\caption{Performance comparison on the \textbf{Ubu-Nav} offline navigation dataset. Coord actions include \texttt{Click}, \texttt{Drag}, \texttt{Scroll}, \texttt{LeftDouble}, and \texttt{RightSingle}; Non-coord actions include \texttt{Type}, \texttt{Hotkey}, \texttt{PressEnter}, and \texttt{Finish}.}
\label{tab:benchmark_pc}
\resizebox{0.85\textwidth}{!}{
\begin{tabular}{l ccc}
\toprule
\textbf{Model} & \textbf{Coord Actions (\%)} & \textbf{Non-coord Actions (\%)} & \textbf{Average (\%)} \\
\midrule
\rowcolor[gray]{0.95}\multicolumn{4}{l}{\textit{Open-source Models}} \\
UI-TARS-7B-SFT \cite{ui_tars_2025} & 32.8 & 4.6 & 28.9 \\
UI-TARS-1.5-7B \cite{ui-tars-15-seed} & 32.2 & 17.4 & 30.2 \\
OS-Atlas-Pro-7B \cite{wu2024osatlas} & 34.2 & 16.0 & 31.7 \\
Holo2-30B-A3B \cite{hai2025holo2modelfamily} & 52.5 & 34.3 & \underline{50.0} \\
Qwen3-VL-30B-A3B \cite{bai2025qwen3vl} & \underline{54.3} & 7.6 & 47.7 \\
UI-Venus-Navi-72B \cite{gu2025ui} & 45.1 & \underline{40.0} & 44.4 \\
\midrule
\rowcolor[gray]{0.95} \textbf{OmegaUse} & \textbf{57.1} & \textbf{48.6} & \textbf{55.9} \\
\bottomrule
\end{tabular}
}
\end{table}

\textbf{Offline Benchmark.} We further assess the agent's fundamental planning and task decomposition capabilities using the {AndroidControl} dataset, which provides high-level instructions that require significant summarization and reasoning. According to the results in Table 6, {OmegaUse} achieves SOTA performance, securing the first place in both evaluated metrics.

Specifically, {OmegaUse} reaches a Type Accuracy of 87.6\% and a Step Success Rate (SR) of 79.1\%. These scores surpass previous leading models such as UI-Venus-Navi-72B (85.9\% Type Acc. / 77.2\% Step SR) and UI-TARS-72B (85.2\% Type Acc. / 74.7\% Step SR). The superior performance on high-level instructions indicates that {OmegaUse} possesses a more robust internal world model for GUI environments.

\textbf{Online Benchmark.} To evaluate real-time interactive capabilities, we employ the AndroidWorld benchmark, which requires agents to navigate dynamic mobile environments. As shown in Table~\ref{tab:android_world}, \text{OmegaUse} achieves a success rate of \textbf{55.7\%}. Notably, \text{OmegaUse} operates as a streamlined end-to-end agent, relying solely on screenshots without the assistance of external planners or Accessibility (A11y) trees.

Despite using fewer input modalities, \text{OmegaUse} demonstrates competitive performance against several larger-scale open-source models. It outperforms UI-TARS-72B (46.6\%) and Aria-UI (44.8\%), while remaining comparable to the high-parameter GLM-4.5v (57.0\%). While a performance gap remains compared to state-of-the-art models such as UI-Venus-Navi-72B (65.9\%), it is worth noting that UI-Venus-Navi-72B is a dense model with a much larger parameter size, whereas OmegaUse is a MoE-based model with a smaller overall parameter size.

\subsubsection{Specialized Offline Benchmarks}

To further evaluate the agent’s generalization across diverse platforms and complex real-world workflows, we conduct experiments on our specialized \textbf{OS-Nav} offline benchmarks: \textbf{ChiM-Nav} for the Chinese mobile ecosystem and \textbf{Ubu-Nav} for Ubuntu desktop environments.

\textbf{ChiM-Nav (Mobile).} This benchmark specifically targets the unique UI layouts and multi-step workflows found in popular applications within the Chinese mobile ecosystem. As shown in Table~\ref{tab:ChiM-Nav}, \text{OmegaUse} achieves a Type Accuracy of \textbf{87.78\%} and a Step Success Rate (SR) of \textbf{74.24\%}, outperforming all existing open-source baselines.

Notably, it surpasses the high-parameter \text{UI-Venus-72b}, which scores 81.23\% Type Acc. and 67.51\% Step SR. The significant lead in Step SR (a gain of approximately 6.7\%) suggests that \text{OmegaUse} is more capable of maintaining reasoning consistency in this scene.

\textbf{Ubu-Nav (Desktop).} The Ubu-Nav benchmark evaluates the agent's proficiency in handling routine Ubuntu desktop operations across varied system interfaces. According to Table~\ref{tab:benchmark_pc}, \text{OmegaUse} reaches an average performance of \textbf{55.9\%}, establishing a clear lead over the best-performing baseline, \text{Holo2-30B-A3B} (50.0\%).

A breakdown of action types reveals that \text{OmegaUse} excels in both coordinate-based actions (\texttt{Click}, \texttt{Drag}, etc.) and non-coordinate actions (\texttt{Type}, \texttt{Hotkey}, etc.). Specifically, it achieves \textbf{48.6\%} in non-coordinate tasks, a substantial improvement over \text{UI-Venus-Navi-72B} (40.0\%) and \text{Holo2-30B-A3B} (34.3\%). These results demonstrate that \text{OmegaUse} effectively bridges the gap between spatial perception and semantic command execution, even in complex desktop environments requiring multi-window coordination.




\section{Conclusion}

In this report, we presented \textbf{OmegaUse}, a high-performance autonomous GUI agent model  capable of navigating complex tasks across mobile and desktop, supporting phone-use and computer-use scenarios. By adopting a Mixture-of-Experts (MoE) backbone, we demonstrate that OmegaUse can maintain superior reasoning depth while significantly optimizing computational efficiency compared to dense models. To build a reliable data foundation, we introduced a carefully engineered data-construction pipeline that combines rigorously curated open-source datasets with an automated synthesis framework integrating bottom-up autonomous exploration and top-down taxonomy-guided generation, thereby producing high-fidelity training trajectories.
To effectively leverage this curated data, we proposed a decoupled two-stage training paradigm, combining SFT with GRPO, successfully calibrates the model's spatial grounding and sequential planning through specialized reward mechanisms. Empirical results across multiple platforms validate the robustness of our approach. OmegaUse establishes new performance records on major benchmarks, notably achieving a SOTA score of \textbf{96.3\%} on ScreenSpot-V2 and a leading \textbf{79.1\%} Step success rate on AndroidControl. Furthermore, we introduce OS-Nav, an offline benchmark for real-world GUI navigation, to enable systematic evaluation of GUI agents in an offline setting. In particular, ChiM-Nav, a Chinese GUI offline benchmark, provides the community with a comprehensive evaluation suite to help bridge the assessment gap within the Chinese digital ecosystem. Additionally, Ubu-Nav is the first offline benchmark designed to evaluate computer-use agents on Ubuntu desktop workflows. Moving forward, we aim to extend OmegaUse’s capabilities to even more intricate, real-world workflows and explore more advanced safety constraints and self-correction mechanisms to ensure reliable and trustworthy autonomous GUI interaction.


\bibliographystyle{unsrt}
\bibliography{refs}




\end{document}